\documentclass[journal]{IEEEtran}

\usepackage{cite}
\usepackage{amsmath,amssymb}
\usepackage{graphicx}
\usepackage{siunitx}
\usepackage{booktabs}
\usepackage{url}
\usepackage[hidelinks]{hyperref}
\usepackage{balance}

\usepackage{stfloats}

\usepackage{float}
\graphicspath{{../}}

\title{Real-Time Semantic Segmentation on FPGA for Autonomous Vehicles Using LMIINet with the CGRA4ML Framework}

\author{Amir~Mohammad~Khadem~Hosseini\ and~Sattar~Mirzakuchaki
\thanks{Corresponding author: Amir Mohammad Khadem Hosseini.}
\thanks{Amir Mohammad Khadem Hosseini and Sattar Mirzakuchaki are with the Department of Electrical Engineering, Iran University of Science and Technology (IUST), Tehran 16846-13114, Iran (e-mails: am\_khadem@cmps2.iust.ac.ir, m\_kuchaki@iust.ac.ir).}}

\begin{document}

\markboth{IEEE Transactions on Circuits and Systems for Artificial Intelligence}{Real-Time Semantic Segmentation on FPGA Using LMIINet}
\maketitle

\begin{abstract}
Semantic segmentation has emerged as a fundamental problem in computer vision, gaining particular importance in real-time applications such as autonomous driving. The main challenge is achieving high accuracy while operating under computational and hardware constraints. In this research, we present an FPGA-based implementation of real-time semantic segmentation leveraging the lightweight LMIINet architecture and the Coarse-Grained Reconfigurable Array for Machine Learning (CGRA4ML) hardware framework. The model was trained using Quantization-Aware Training (QAT) with 8-bit precision on the Cityscapes dataset, reducing memory footprint by a factor of four while enabling efficient fixed-point computations. Necessary modifications were applied to adapt the model to CGRA4ML constraints, including simplifying skip connections, employing hardware-friendly operations such as depthwise-separable and 1A-1 convolutions, and redesigning parts of the Flatten Transformer. Our implementation achieves approximately 90\% pixel accuracy and 45\% mean Intersection-over-Union (mIoU), operating in real-time at \SI{20}{frames~per~second~(FPS)} with \SI{50.1}{ms} latency on the ZCU104 FPGA board. The results demonstrate the potential of CGRA4ML, with its flexibility in mapping modern layers and off-chip memory utilization for skip connections, provides a path for implementing advanced semantic segmentation networks on FPGA for real-time applications to outperform traditional GPU solutions in terms of power efficiency while maintaining competitive accuracy. The code for this project is publicly available at \url{https://github.com/STAmirr/cgra4ml_semantic_segmentation}.
\end{abstract}


\section{Introduction}
\IEEEPARstart{R}{eliable} autonomous driving demands dense scene understanding at video rates under strict power, thermal, and cost constraints. Among perception tasks, \emph{semantic segmentation} provides pixel-wise interpretation of a scene---separating roadway, sidewalk, vehicles, pedestrians, and signage---and directly supports planning and control. Over the last decade, accuracy has improved rapidly with deeper backbones and attention/Transformer modules; however, typical GPU deployments remain power-hungry and introduce non-deterministic latency due to driver scheduling, memory traffic, and batching requirements. These properties are problematic for embedded electronic control units (ECUs) that must meet hard end-to-end deadlines and functional-safety constraints in automotive environments.

\subsection{Why FPGA for real-time semantic segmentation.}
Field-programmable gate arrays (FPGAs) provide fine-grained spatial parallelism and deterministic, clocked pipelines with tight control over the memory hierarchy. Compared to general-purpose GPUs, the accelerator datapath can be specialized to the exact operations in the network (e.g., depthwise separable convolutions, pointwise convolutions, attention projections), enabling:
\begin{itemize}
  \item \textbf{Deterministic latency:} deep pipelines with static initiation intervals (II) and bounded queues, avoiding OS and driver jitter. End-to-end latency scales with pipeline depth rather than input batch size; real-time inference at batch $b{=}1$ is natural rather than penalized.
  \item \textbf{Reduced data movement:} line buffers and on-chip SRAM tile the feature maps to minimize off-chip DRAM traffic---a dominant source of latency and energy on GPUs. Streaming between layers eliminates intermediate host-device transfers.
  \item \textbf{Quantization-first arithmetic:} fixed-point INT8/INT4 MAC arrays match the needs of quantization-aware training (QAT), lowering energy per operation while preserving accuracy under careful calibration.
  \item \textbf{Operator fusion and dataflow execution:} custom pipelines fuse convolution, normalization, and activation, collapsing kernel boundaries and removing launch overheads.
  \item \textbf{Safety and lifecycle:} reconfigurability enables over-the-air updates and field patches while preserving a verified timing budget---useful for ISO~26262 processes.
\end{itemize}

\subsection{From FPGA to ASIC: The Migration Path and Cost Rationale}
A hardware-software co-design validated on FPGA establishes the micro-architecture (compute arrays, buffer sizes, dataflow, and quantization) that can later be hardened as an ASIC. While ASICs incur non-recurring engineering (NRE) costs, at production scale their unit cost and energy per inference are substantially lower than GPU modules or large FPGAs. Crucially, the design techniques we employ---tiling sizes, reuse factors, loop unrolling, operand bit-widths, and on-chip buffer topology---transfer directly to an ASIC netlist. Thus, the same design that yields deterministic $\leq\SI{50}{\milli\second}$ latency and real-time throughput on FPGA provides a blueprint for a cost-optimized ASIC with even lower latency, higher TOPS/W, and reduced bill-of-materials at volume.

\subsection{Problem setting and gap.}
Lightweight convolutional networks such as ENet and ERFNet \cite{ref5,ref6} and more recent hybrid CNN-Transformer models (e.g., SegFormer and LMIINet \cite{ref7,ref1}) deliver compelling accuracy-speed trade-offs on GPUs. FPGA implementations to date have typically focused on very compact CNNs or aggressive quantization (e.g., ENetHQ) to reach real-time, often sacrificing accuracy on benchmarks like Cityscapes \cite{ref9}. Conversely, Transformer-augmented models improve global context reasoning but are rarely mapped efficiently to reconfigurable logic due to memory access patterns and attention bottlenecks. This work bridges that gap via a co-design that preserves accuracy-critical components of LMIINet while restructuring the network and its schedule for a spatial dataflow backend.

\subsection{Our approach in brief.}
We re-architect \emph{LMIINet} for FPGA deployment and map it with the \emph{CGRA4ML} framework. The network integrates a lightweight encoder-decoder with a reduced \emph{Flatten Transformer} block and channel attention modules; we apply QAT at \SI{8}{bit} to align with integer arithmetic on the fabric. CGRA4ML generates SystemVerilog for a coarse-grained reconfigurable array (CGRA) that implements the convolutional and attention operators as streaming kernels with line-buffered feature tiles and double-buffered weight blocks. The flow emphasizes (i) keeping activations on-chip, (ii) minimizing off-chip transfers, (iii) fusing operators when possible, and (iv) exploiting depthwise/pointwise separability to maximize reuse.

\subsection{Objectives.}
Our design targets real-time perception on the Xilinx ZCU104 while preserving segmentation quality on Cityscapes:
\begin{itemize}
  \item \textbf{Latency/throughput:} $\leq\SI{50}{ms}$ end-to-end latency at $\geq\SI{20}{FPS}$ for batch $b{=}1$.
  \item \textbf{Accuracy:} $\sim 90\%$ pixel accuracy and $\sim 45\%$ mIoU on Cityscapes validation \cite{ref9} with 19 evaluation classes.
  \item \textbf{Resource efficiency:} bounded LUT/FF/BRAM usage compatible with mid-range automotive-grade FPGAs; a dataflow that is directly portable to ASIC.
\end{itemize}

\subsection{Contributions.}
This paper makes the following contributions:
\begin{itemize}
  \item A quantized, FPGA-compatible LMIINet variant that retains global-context modeling with a hardware-friendly \emph{Flatten Transformer} and channel attention.
  \item A CGRA4ML-based mapping that realizes operator fusion, streaming dataflow, and on-chip buffering for deterministic latency at batch $b{=}1$.
  \item A four-phase QAT schedule and measurement protocol that disentangles accuracy from pure throughput, enabling fair comparison to ENetHQ (FPGA) and GPU baselines.
  \item An end-to-end implementation on ZCU104 demonstrating real-time performance with competitive mIoU on Cityscapes and a discussion of the migration path to ASIC for reduced unit cost and energy.
\end{itemize}

Experiments are conducted on the Cityscapes dataset \cite{ref9}, a widely used benchmark comprising high-resolution images (2048$\times$1024) captured across 50 European cities with fine-grained annotations for 19 classes. The remainder of the paper is organized as follows. Section~2 reviews related work on real-time segmentation and FPGA acceleration. Section~3 details the \emph{Materials and Methods}: LMIINet architecture, CGRA4ML mapping, dataset and preprocessing, quantization-aware training schedule, and measurement protocol. Section~4 reports results and comparisons with GPU and ENetHQ baselines. Section~5 concludes the paper and Section~6 outlines future directions.

\section{Related Work}

\subsection{Real-Time Semantic Segmentation Networks}
Real-time semantic segmentation for autonomous driving has driven the development of efficient deep neural network architectures. Early lightweight CNN models such as ENet \cite{ref5} and ERFNet \cite{ref6} prioritized low latency, achieving moderate accuracy on Cityscapes at high frame rates. For example, ENet contains an extremely compact encoder-decoder design (0.37~million parameters) that runs over \SI{100}{FPS} on high-end GPUs but attains only around 58--65\% mean IoU on Cityscapes. ERFNet improved this trade-off by introducing residual factorized convolutions, reaching about 72.5\% mIoU with \SI{7}{FPS} on a Jetson~TX1 (and \SI{83}{FPS} on a Titan~X). Subsequent designs like BiSeNet employed two-path architectures to preserve spatial detail alongside context, yielding 68.4\% mIoU on the Cityscapes test dataset at \SI{105}{FPS} on one NVIDIA Titan~XP \cite{ref7}. More recent approaches integrate Transformer-based modules to capture global context. SegFormer \cite{ref8} exemplifies a pure Transformer encoder that achieves state-of-the-art accuracy (over 80\% mIoU on Cityscapes) by using multi-scale attention, albeit with high computational cost. Hybrid architectures have emerged to balance these aspects; LMIINet \cite{ref1} is one such model that combines efficient CNN bottlenecks with a lightweight Flatten Transformer, attaining 72.0\% mIoU at \SI{100}{FPS} on Cityscapes (RTX~2080Ti GPU) \cite{ref1}. These methods demonstrate the progression from purely convolutional networks to CNN+Transformer hybrids in pursuit of better accuracy--speed trade-offs for autonomous driving.

\subsection{FPGA-Based Acceleration and CGRA Frameworks}
Deploying semantic segmentation on FPGAs has attracted attention for its potential to meet the strict power and latency requirements of edge automotive systems \cite{ref3}. However, implementing deep networks on reconfigurable logic is challenging due to limited on-chip resources and memory bandwidth. The work of Ghielmetti \textit{et al.} \cite{ref2} (using the \textit{hls4ml} toolkit) demonstrated a fully on-chip FPGA implementation of a highly pruned and quantized ENet model. Their heterogeneous-quantized ENetHQ network (with layer-wise mixed precision) achieved ${\sim}81\%$ pixel accuracy and 36.8\% mIoU on Cityscapes, with a 4.9~ms latency per image on a Xilinx ZCU102 board \cite{ref2}. This design used only about 25--30\% of the FPGA's resources by aggressively reducing filter counts and precisions \cite{ref2}, indicating the power savings possible with tailored networks (the batch-10 latency was further reduced to 3~ms at the cost of buffering delay) \cite{ref2}. Other FPGA implementations include Jia \textit{et al.}'s work deploying the original ENet on a Zynq-7035 via Xilinx's DPU, though with higher latency ($\sim$30~ms) due to external memory transfers. In general, prior FPGA solutions have traded off accuracy for speed or vice versa, as most high-accuracy models could not fit fully on-chip \cite{ref4}. 

To address the limitations of HLS-based workflows like hls4ml (which require storing all weights on chip), the Coarse-Grained Reconfigurable Array for Machine Learning (CGRA4ML) framework was proposed \cite{ref4}. CGRA4ML employs a coarse-grained reconfigurable array that allows layers to stream data off-chip, enabling support for larger and more complex networks that are beyond the reach of pure HLS designs \cite{ref4}. Notably, CGRA4ML generates synthesizable SystemVerilog for flexible deployment on FPGA or ASIC, and provides a runtime for dynamic reconfiguration \cite{ref4}. Our work builds upon these advances by leveraging CGRA4ML to implement a modern segmentation model. We co-designed the LMIINet architecture with hardware constraints in mind, simplifying unsupported operations and using 8-bit quantization throughout, similar to Ghielmetti \textit{et al.}'s quantization-aware training approach \cite{ref2}. Unlike the ENetHQ design which sacrifices mIoU for extreme speed, our CGRA4ML-based LMIINet maintains a higher accuracy (45\% mIoU) while still operating in real time (20~FPS) on the ZCU104 FPGA. This is achieved through a hardware--software co-design: the network architecture was tailored to CGRA4ML's dataflow (e.g., replacing complex skip connections with memory-friendly ones) and the hardware was configured (16$\times$96 PEs, \SI{200}{MHz} clock) to optimally map the modified LMIINet. Overall, the proposed approach improves upon prior art by delivering a better balance of accuracy, latency, and power efficiency in semantic segmentation on FPGAs, demonstrating that even advanced CNN--Transformer hybrids can be deployed under strict resource constraints.

\section{Materials and Methods}

\subsection{Model Architecture (LMIINet)}
LMIINet \cite{ref1} is a lightweight semantic segmentation model designed to balance accuracy and computational efficiency, particularly in real-time autonomous driving applications. It integrates CNN-based feature extraction with Transformer-based global context modeling to capture both local spatial features and global dependencies in the image.

For CGRA4ML deployment, several architectural modifications were implemented to ensure compatibility with the framework's constraints and optimize hardware mapping efficiency.

\begin{figure*}[!t]
    \centering
    \includegraphics[width=\textwidth]{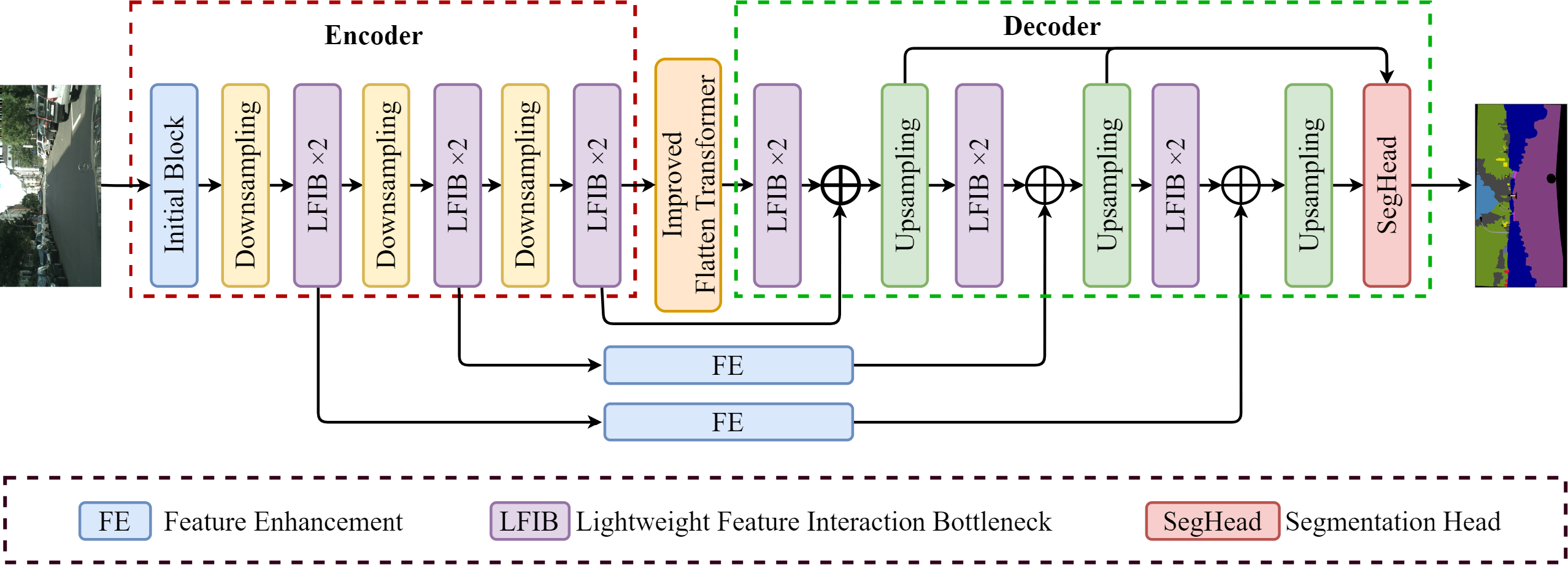}
    \caption{LMIINet Architecture \cite{ref1}}
    \label{fig:lminet_architecture}
\end{figure*}

\subsubsection{Encoder-Decoder Structure}
LMIINet uses an encoder-decoder architecture similar to U-Net, but with optimizations to reduce computational complexity. Encoder extracts local features from the image using depthwise separable convolutions and asymmetric convolutions. Decoder restores spatial resolution using nearest-neighbor upsampling and depthwise convolutions to ensure fine-grained segmentation.

\textbf{CGRA4ML Modifications:}
- Added initial downsampling stem with 3×3 convolution and stride-2 for efficient feature extraction
- Implemented progressive encoder blocks with filters [24, 48, 96, 128] instead of the standard configuration
- Modified decoder to use nearest-neighbor upsampling exclusively for CGRA4ML compatibility
- Added spatial refinement using depthwise-pointwise convolution pairs to smooth upsampling artifacts

\begin{figure*}[!t]
    \centering
    \includegraphics[width=\textwidth]{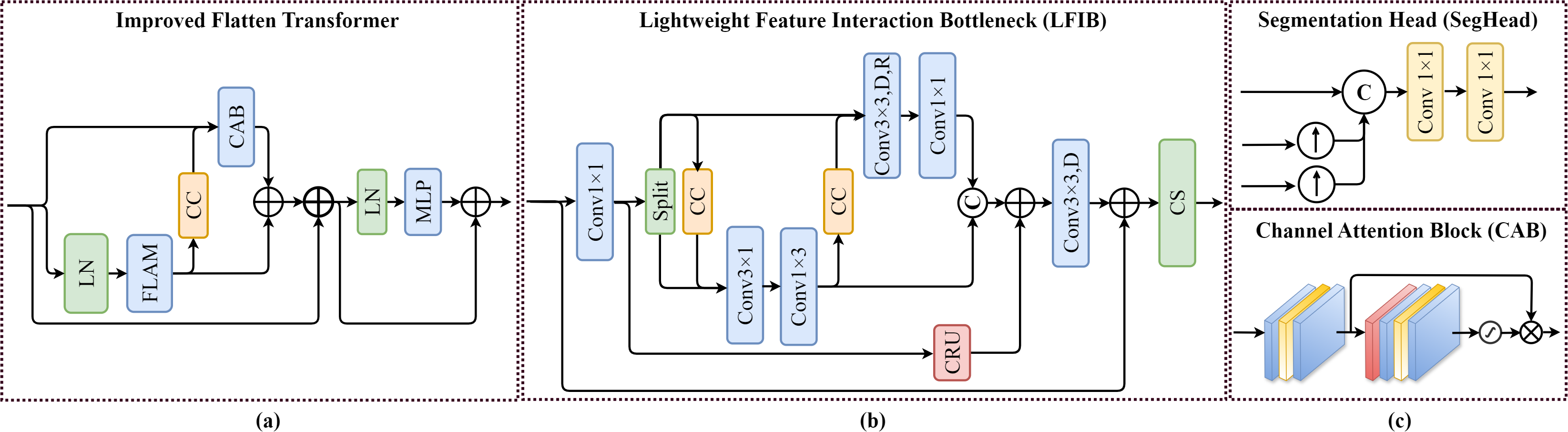}
    \caption{The diagram of the proposed Lightweight Feature Interaction Bottleneck (LFIB), improved Flatten Transformer, Segmentation Head (SegHead), and
Channel Attention Block (CAB). D represents the depth-wise convolution, R is the kernel of dilated convolution, and CS denotes the channel shuffle operation. \cite{ref1}}
    \label{fig:lmiinet2}
\end{figure*}

\begin{figure*}[!t]
    \centering
    \includegraphics[width=\textwidth]{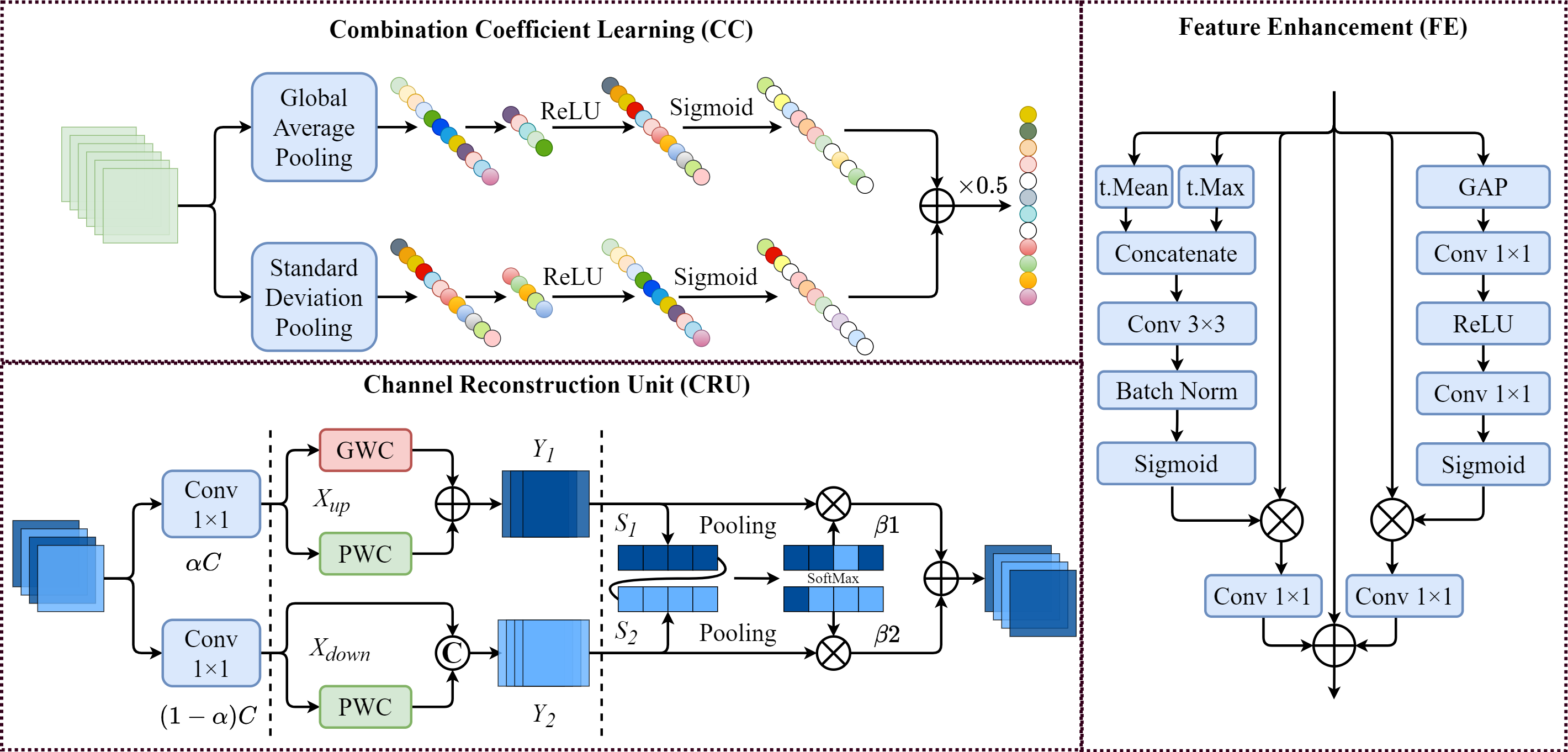}
    \caption{The diagram of the Combination Coefficient learning (CC) scheme, Channel Recurrent Unit (CRU), and the Feature Enhancement (FE) module \cite{ref1}}
    \label{fig:lmiinet3}
\end{figure*}

\subsubsection{Lightweight Feature Interaction Bottleneck (LFIB)}
The LFIB module is at the core of LMIINet, designed to minimize computational load while improving feature interaction:
\begin{itemize}
\item Depthwise Separable Convolutions reduce the number of parameters and computational complexity.
\item Asymmetric Convolutions (e.g., 3×1 and 1×3) allow for flexible receptive field adjustments.
\item Dilated Convolutions expand the receptive field without increasing computational cost.
\end{itemize}

\textbf{CGRA4ML Modifications:}
- Implemented simplified channel splitting using tensor slicing operations for hardware efficiency
- Added Combination Coefficient (CC) Figure~\ref{fig:lmiinet3} modules using Global Average Pooling and Dense layers for channel-wise attention
- Integrated Channel Reconstruction Unit (CRU) blocks for refined feature processing
- Replaced complex branching with streamlined left-right branch architecture using asymmetric convolutions and depthwise separable convolutions
- Added CGRA4ML-compatible channel shuffle using concatenation operations instead of complex permutations

\subsubsection{Flatten Transformer}
The Flatten Transformer improves the integration of local spatial features and global contextual information:
- The transformer enhances global context modeling by using a focused linear attention module (FLAM)  to reduce computational complexity.
- The integration of local spatial features ensures that fine details are preserved during segmentation.

\textbf{CGRA4ML Modifications:}
- Replaced complex multi-head attention with simplified Flattened Local Attention Module (FLAM) using dilated convolutions
- Implemented CGRA4ML-compatible attention using 3×3 convolutions with dilation rate 2 instead of full transformer blocks
- Reduced transformer heads from 16 to 8 and key dimensions from 128 to 64 for hardware efficiency
- Added residual connections and batch normalization for stable gradient flow

\subsubsection{Channel Attention Block (CAB)}
The CAB illustrated in Figure~\ref{fig:lmiinet2} (c) is used to highlight important channels by applying an attention mechanism. It enables the model to focus on key features while suppressing irrelevant ones, improving the model's performance.

\textbf{CGRA4ML Modifications:}
- Simplified CAB implementation using Global Average Pooling followed by two 1×1 convolutions
- Replaced complex attention mechanisms with sigmoid activation for channel-wise scaling
- Added Feature Enhancement (FE) Figure~\ref{fig:lmiinet3} blocks combining Global Average Pooling and Global Max Pooling for better context aggregation
- Implemented multiply operations for efficient channel attention without complex branching

\subsubsection{Segmentation Head}
The segmentation head illustrated in Figure~\ref{fig:lmiinet2} (c) generates the final output by combining features from different scales. Multi-scale fusion allows the model to recover fine-grained details and global context, ensuring high segmentation accuracy.

\textbf{CGRA4ML Modifications:}
- Implemented multi-scale feature fusion using only upsampling and concatenation operations
- Added auxiliary segmentation head at 1/8 scale for improved training convergence
- Simplified final classification using single 1×1 convolution followed by softmax activation
- Optimized for CGRA4ML's dataflow architecture with reduced memory access patterns
- Used only the last three decoder features for segmentation to reduce computational overhead

\subsubsection{CGRA4ML Mapping and Hardware Configuration}
the Coarse-Grained Reconfigurable Array for Machine Learning (CGRA4ML) framework is designed to efficiently implement neural networks on FPGA. Unlike traditional HLS4ML, which struggles with larger models, CGRA4ML offers dynamic reconfiguration and supports off-chip data storage, making it suitable for larger, more complex models.

\subsubsection{CGRA4ML Workflow}
The CGRA4ML workflow involves the following steps:
\begin{enumerate}
    \item \textbf{Model Definition}: Train the LMIINet model with Quantization-Aware Training (QAT) to reduce the computational burden.
    \item \textbf{Hardware Mapping}: The trained model is converted into SystemVerilog RTL using the CGRA4ML toolchain.
    \item \textbf{Simulation}: Verify the hardware implementation using Verilator and other simulators.
    \item \textbf{FPGA Synthesis}: The final design is synthesized on the ZCU104 FPGA platform.
    \item \textbf{Real-Time Inference}: The model is deployed on FPGA for real-time semantic segmentation.
\end{enumerate}

\begin{figure*}[!t]
    \centering
    \includegraphics[width=\textwidth]{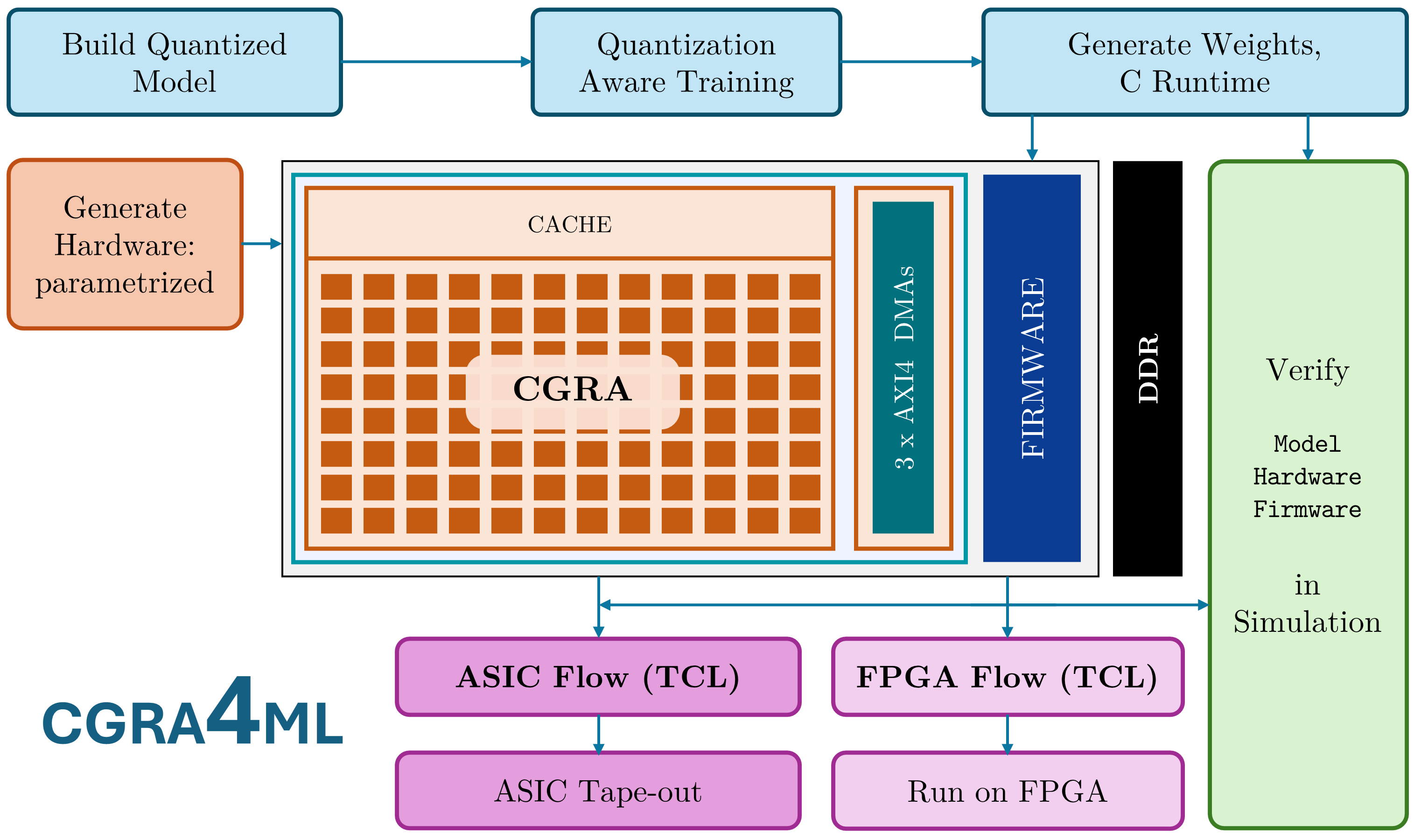}
    \caption{CGRA4ML FPGA Implementation \cite{ref4}}
    \label{fig:cgra4ml_architecture}
\end{figure*}

\subsubsection{Hardware Configuration}
The CGRA4ML hardware configuration is optimized for real-time semantic segmentation, balancing computational throughput, memory efficiency, and power consumption. The key parameters are organized into four main categories:

\begin{itemize}
    \item \textbf{Computational Resources}: $(16 \times 96)$ PE array providing 1,536 parallel processing units, \SI{200}{MHz} clock frequency for \SI{5}{ns} cycle time, and maximum batch size of 2 images for simultaneous processing and throughput optimization.
    
    \item \textbf{Data Precision and Quantization}: 8-bit precision for input activations and weights reducing memory footprint by 4$\times$ compared to FP32, 32-bit accumulation precision for multiply-accumulate operations preventing overflow, and 16-bit bias precision providing sufficient dynamic range.
    
    \item \textbf{Memory Architecture}: 256-depth RAM for on-chip weight storage and local caching, 32,768-depth RAM for edge data storage supporting intermediate feature maps and skip connections, and support for up to 256 input channels enabling wide feature maps in deep networks.
    
    \item \textbf{Data Interface and Communication}: 128-bit AXI bus width for high-bandwidth memory transfers, maximum burst length of 16 beats for optimized memory access efficiency, 64-bit header width for streaming data packet management, and support for up to $9 \times 9$ kernel sizes enabling large receptive fields and dilated convolutions.
\end{itemize}

This configuration enables efficient mapping of the modified LMIINet architecture while maintaining real-time performance constraints for autonomous driving applications.

\subsection{Dataset and Preprocessing}

We use the Cityscapes benchmark for semantic urban scene understanding, which provides \emph{5,000} finely annotated images (2,975 train / 500 val / 1,525 test) and an additional \emph{20,000} coarsely annotated images captured across 50 European cities. Images are 2048$\times$1024 resolution with 19 evaluation classes drawn from eight categories (e.g., flat, construction, nature, vehicle, sky, object, human, void). The dataset design emphasizes diverse, complex inner-city scenes and balanced train/val/test splits by city and season \cite{ref9}. In all experiments, we follow the common practice of training on fine annotations (train) and evaluating on the validation split; coarse annotations can optionally be used for pretraining or auxiliary supervision. Standard photometric and geometric augmentations (random cropping, horizontal flipping, color jitter) are applied, along with task-specific zoom-to-signs augmentation described later.

\subsection{Quantization-Aware Training and Schedule}

The quantized LMIINet was trained on the Cityscapes dataset using Quantization-Aware Training (QAT) with 8-bit activations and weights to ensure FPGA compatibility. 
Training spanned 240 epochs with a batch size of 2 and an initial learning rate of $7\times10^{-4}$ using stochastic gradient descent (momentum 0.9, Nesterov).
  
\begin{figure*}[!t]
    \centering
    \includegraphics[width=\textwidth]{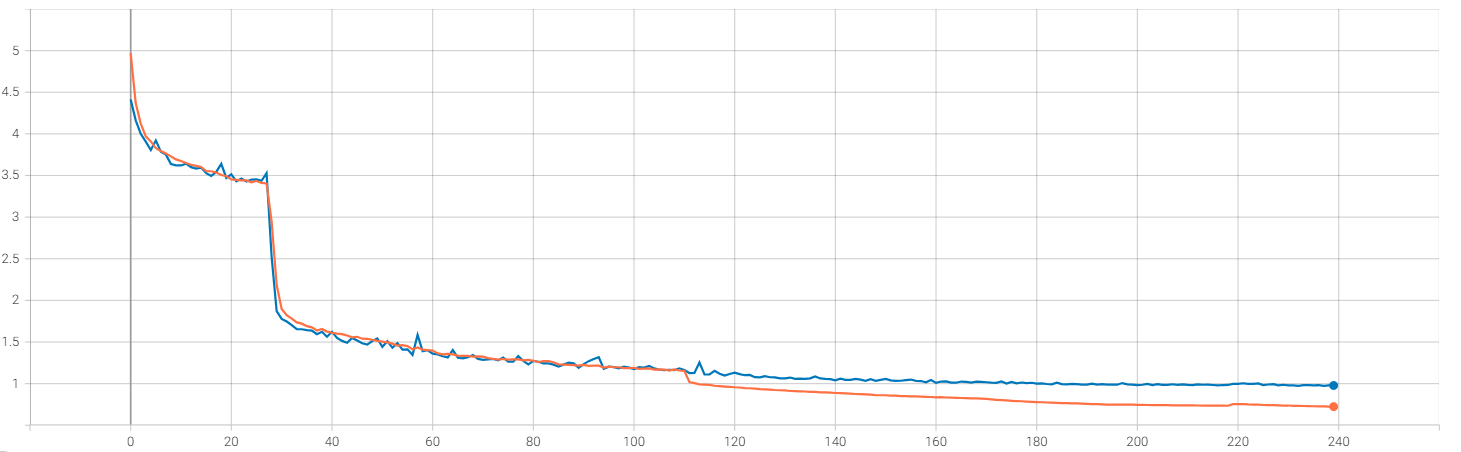}
    \caption{Training (orange) and validation (blue) loss curves for quantized LMIINet over 218 epochs. Loss decreases sharply in Phase~1, stabilizes during Phase~2 (with frozen decoder), then drops further in Phase~3 after regularization removal, and finally levels off in Phase~4.}
    \label{fig:loss_curve}
\end{figure*}

\begin{figure*}[!t]
    \centering
    \includegraphics[width=\textwidth]{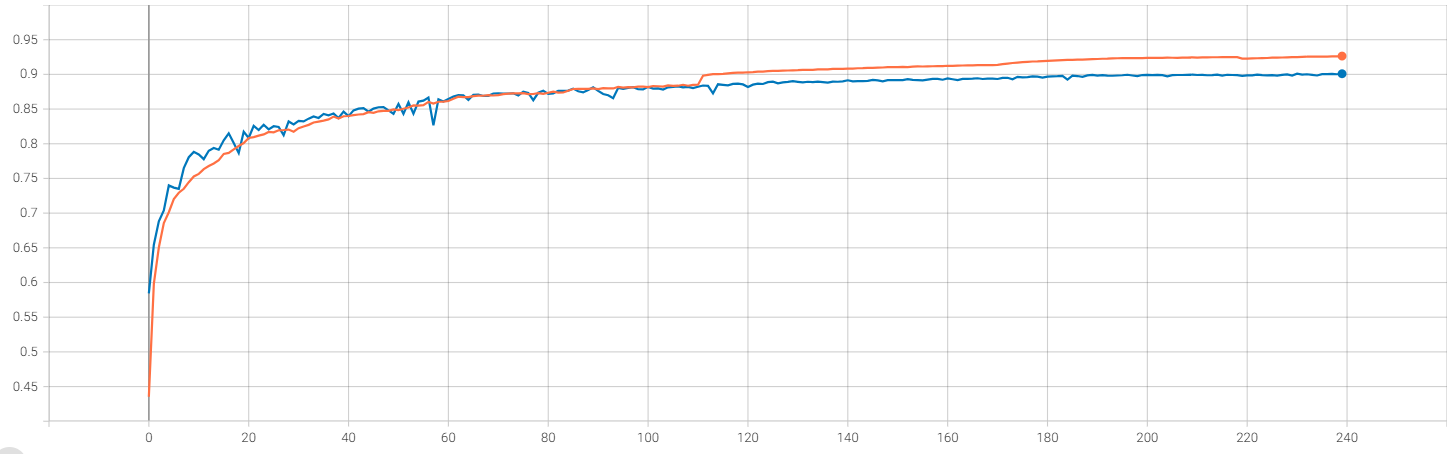}
    \caption{Pixel accuracy curves for training (orange) and validation (blue). The model reaches about 90\% pixel-wise accuracy. Notable jumps correspond to Phase~2 (frozen decoder stability) and Phase~3 (fine-tuning without augmentation).}
    \label{fig:acc_curve}
\end{figure*}

\begin{figure*}[!t]
    \centering
    \includegraphics[width=\textwidth]{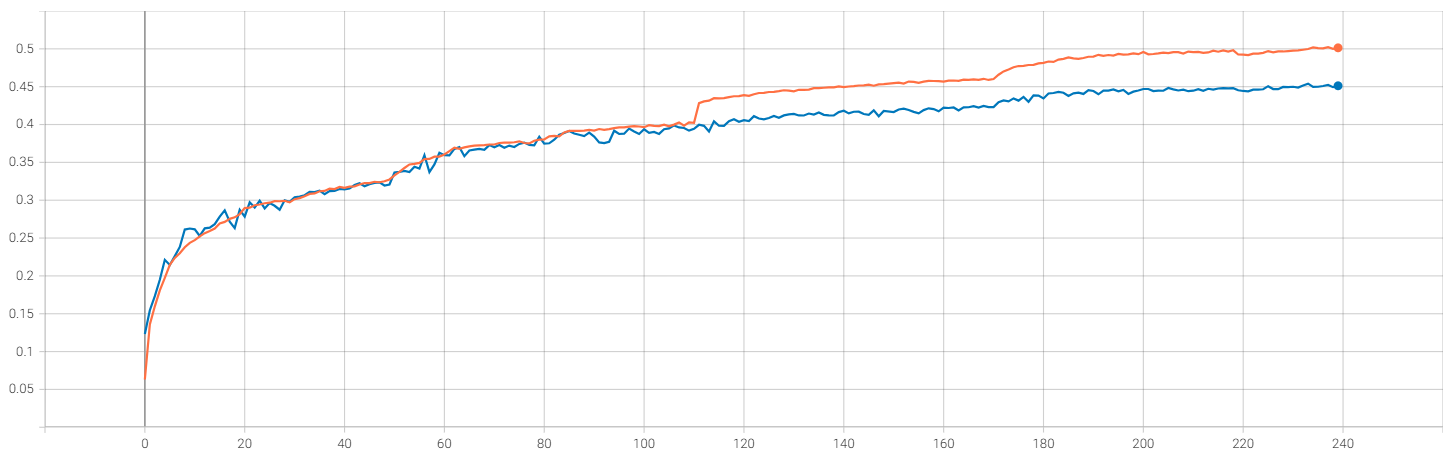}
    \caption{Mean Intersection-over-Union (mIoU) curve for training and validation. The quantized model stabilized around 45\% mIoU. X-axis: Epochs, Y-axis: Mean Intersection-over-Union (mIoU). The peak validation mIoU of 0.452 was reached at epoch 239. Orange line represents training mIoU and blue line represents validation mIoU.}
    \label{fig:miou_curve}
\end{figure*}

To stabilize convergence and reduce overfitting, a four-phase schedule was adopted, summarized in Table \ref{tab:training_schedule}.  
During the first 50 epochs, both dropout and data augmentation were enabled while the decoder remained unfrozen and auxiliary (AUX) supervision was disabled. 
This stage emphasized generalization and robust feature extraction, producing the steep initial drop in loss and rapid rise in accuracy visible in Figures \ref{fig:loss_curve} and \ref{fig:acc_curve}. 
Between epochs 50--110, dropout and augmentation remained active while the decoder was frozen and AUX supervision activated, allowing the encoder and transformer blocks to stabilize without decoder noise; this reduced oscillations in validation curves. 
From epochs 110--170, regularization and augmentation were disabled to let the network fit finer spatial details, with the frozen decoder and active AUX head guiding feature alignment, yielding the noticeable step increase in validation accuracy and mIoU around epoch 110. 
Finally, in epochs 170--218, dropout and augmentation were re-enabled and the decoder unfrozen to jointly fine-tune all layers under mild regularization, improving boundary precision and mitigating overfitting before convergence.  
As shown in Figures \ref{fig:loss_curve}--\ref{fig:miou_curve}, these controlled training phases resulted in a stable optimization process.

\begin{table}[t]
\centering
\caption{Training schedule for quantized LMIINet model using Qkeras.}\label{tab:training_schedule}
\begin{tabular}{cccccc}
\toprule
Phase & Epochs & \shortstack{Frozen\\DEC} & \shortstack{AUX\\Supervision} & Dropout & Augmentation \\
\midrule
1 & 0--50   & No  & Off & On  & On  \\
2 & 50--110 & Yes & On  & On  & On  \\
3 & 110--170 & Yes & On  & Off & Off \\
4 & 170--218 & No  & On  & On  & On  \\
\bottomrule
\end{tabular}
\end{table}

The learning rate was decayed by a factor of 0.1 at epochs 110 and 170 to facilitate convergence in the later fine-tuning phases. This multi-stage approach allowed LMIINet to progressively learn robust features and then refine details for segmentation accuracy.

Figure \ref{fig:loss_curve} illustrates the training and validation loss curves, showing an initial rapid decline followed by minor oscillations during the phase transitions, and finally a smooth plateau as training converges. Figure \ref{fig:acc_curve} similarly shows the pixel accuracy rising steadily to around 90\%, while Figure \ref{fig:miou_curve} shows the mean IoU reaching approximately 45\% on the validation set.

\subsection{Measurement Protocol}

We report standard segmentation metrics on Cityscapes, namely pixel accuracy and mean Intersection-over-Union (mIoU) computed on the validation set. For hardware latency, we measure single-image end-to-end latency and throughput (FPS) on the Xilinx ZCU104 at \SI{200}{MHz} using cycle-accurate Verilator simulation of the generated SystemVerilog design; layer-wise cycles and utilization are recorded to derive overall latency. Resource utilization (LUTs, FFs, BRAM) is obtained from synthesis reports. Comparative baselines (GPU and ENetHQ on FPGA) are evaluated with the same dataset and metrics, with details deferred to the Results section.

\section{Results}

\subsection{Training and Model Performance}

At the end of training, the quantized LMIINet achieved approximately 90.0\% pixel accuracy and 45.2\% mIoU on the Cityscapes validation set. These metrics are on par with the unquantized LMIINet model \cite{ref1}, indicating that QAT effectively preserved accuracy despite the use of 8-bit weights and activations.

\subsection{Qualitative Results of LMIINet Predictions}

\begin{figure*}[!t]
    \centering
    \includegraphics[width=\textwidth]{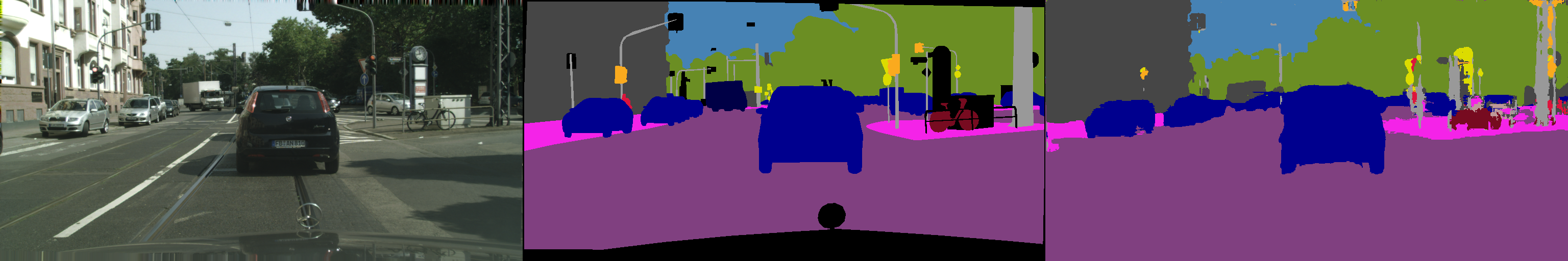}
    \includegraphics[width=\textwidth]{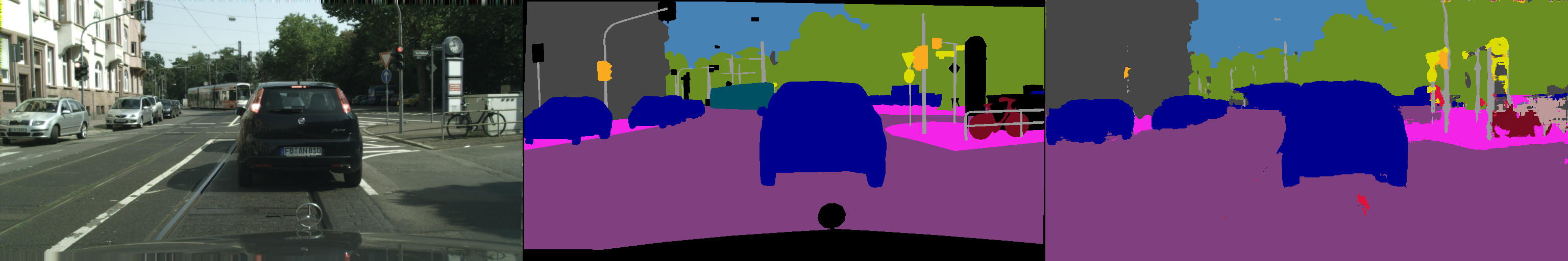}
    \includegraphics[width=\textwidth]{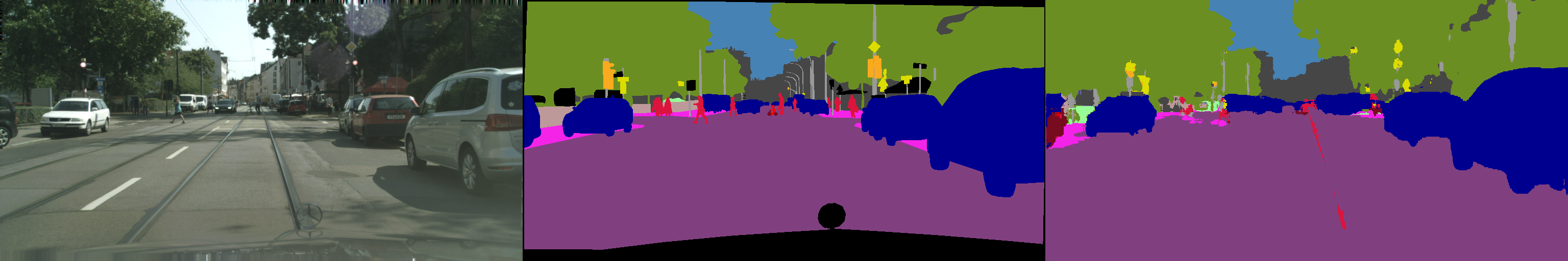}
    \caption{Qualitative results of the modified LMIINet model after 240 epochs of training. For each example: left is the real input image, middle is the ground truth label, and right is the model prediction.}
    \label{fig:keras_outputs}
\end{figure*}

Figure~\ref{fig:keras_outputs} shows qualitative results of the modified LMIINet model after 240 epochs of training. For each example: the left image is the real input image, the middle is the ground truth label, and the right is the model prediction. The model successfully segments critical classes such as road, vehicles, and pedestrians with high fidelity, even under challenging lighting and occlusion conditions.

The qualitative examples highlight LMIINet's ability to capture fine details (e.g., road markings, distant pedestrians) while maintaining global context (segmentation of large road areas and sky). This confirms that the Flatten Transformer module contributes to understanding scene-wide structure, complementing the localized details provided by the CNN encoder-decoder.

\subsection{FPGA Implementation and Performance}

The FPGA implementation achieved:
\begin{itemize}
    \item \textbf{Latency}: \SI{50}{ms} per frame, ensuring real-time processing for autonomous vehicles.
    \item \textbf{Throughput}: The system processed \SI{20}{FPS}, suitable for live video feeds in autonomous driving applications.
\end{itemize}

\subsection{Verilator Simulation Results}

To accurately measure the cycle-by-cycle performance of the LMIINet model on FPGA, we ran a Verilator simulation. The table below shows the number of operations, simulation cycles, utilization, and memory usage for each layer of the network. Multiplying the cycles by the clock period (\SI{5}{ns} for \SI{200}{MHz}) yields an approximate latency of \SI{50}{ms} per frame.

\begin{table}[t]
\centering
\caption{Layer-wise Verilator simulation results for LMIINet on FPGA}
\label{tab:verilator_sim}
\begin{tabular}{cccccc}
\hline
Layer & Ops (M) & Cycles & Util (\%) & Mem (MB) & Type \\
\hline
0 & 339.7 & 327,681 & 67.5 & 13.8 & Conv+Pool \\
1 & 1359.0 & 1,196,034 & 74.0 & 13.1 & Conv+Pool \\
2 & 1359.0 & 890,883 & 99.3 & 8.4 & Conv+Pool \\
3 & 906.0 & 593,928 & 99.3 & 5.6 & Conv+Pool \\
4 & 302.0 & 197,640 & 99.5 & 1.8 & Conv+Pool \\
5 & 33.6 & 33,026 & 66.1 & 0.7 & Conv+Pool \\
6 & 302.0 & 197,640 & 99.5 & 1.8 & Conv+Pool \\
7 & 33.6 & 32,897 & 66.4 & 0.7 & Conv+Pool \\
8 & 906.0 & 592,902 & 99.5 & 5.1 & Upsample \\
9 & 75.5 & 131,585 & 37.4 & 2.6 & Conv \\
10 & 1359.0 & 1,187,844 & 74.5 & 10.1 & Conv+Pool \\
11 & 75.5 & 526,337 & 9.3 & 9.0 & Conv \\
12 & 1359.0 & 1,187,841 & 74.5 & 10.1 & Conv+Pool \\
13 & 75.5 & 2,105,345 & 2.3 & 33.0 & Conv \\
14 & 239.1 & 819,201 & 19.0 & 20.8 & Conv+Softmax \\
\hline
\end{tabular}
\end{table}

The overall simulation confirms a latency of approximately 50 ms per frame at 200 MHz. Utilization peaks around 99\% for several convolutional layers, indicating efficient PE usage for heavy layers, while lighter layers maintain lower utilization.

\subsection{Comparison with GPU Solutions}
When compared to GPU-based solutions, the FPGA-based LMIINet implementation demonstrates a significant reduction in power consumption while maintaining competitive performance in terms of accuracy and latency. The ENetHQ model, although faster in terms of FPS, has a considerably lower mIoU (36.8\%) compared to the 45\% mIoU of LMIINet.

\begin{table*}[t]
\centering
\caption{Performance Comparison of Final Models on GPU and FPGA}
\label{tab:performance_comparison}
\begin{tabular*}{\textwidth}{@{\extracolsep{\fill}}lccc@{}}
\toprule
\textbf{Metric} & \textbf{GPU (2080Ti)} & \textbf{ENetHQ FPGA \cite{ref2}} & \textbf{LMIINet FPGA (Ours)} \\
\midrule
\textbf{Pixel Accuracy}          & 90.0\% & 81.1\% & 90.0\% \\
\textbf{mIoU}                    & 45.0\% & 36.8\% & 45.0\% \\
\textbf{Latency (per frame)}    & \SI{19.57}{ms} & \SI{4.9}{ms} (b1) / \SI{3.06}{ms} (b10) & \SI{50.10}{ms} \\
\textbf{FPS}                     & \SI{51.1}{FPS} & \SI{32.7}{FPS} (b1) / \SI{327}{FPS} (b10)$^{*}$ & \SI{19.95}{FPS} \\
\textbf{BRAM Utilization}       & --- & 224.5 (25\%) & 6.0 (1.92\%) \\
\textbf{LUT Utilization}        & --- & 342.0 (37\%) & 206,830 (89.77\%) \\
\textbf{FF Utilization}         & --- & 87,059 (16\%) & 196,980 (42.75\%) \\
\bottomrule
\end{tabular*}
\end{table*}

In Table~\ref{tab:performance_comparison}, we compare the performance of the final models on GPU, the ENetHQ model \cite{ref2}, and our FPGA model. Key metrics include pixel accuracy, mean intersection over union (mIoU), latency per frame, frames per second (FPS), and hardware utilization (BRAM, LUT, FF usage). The ENetHQ design achieves extremely low latency by running a simplified network fully on-chip, but at the cost of lower mIoU. Our FPGA implementation, while slower in absolute FPS, delivers significantly higher accuracy (matching the GPU results) and remains within real-time requirements. Notably, the ENetHQ design utilized only 25--37\% of the ZCU102's resources due to its compact architecture, whereas our LMIINet design utilizes a larger portion of the ZCU104 resources to accommodate the more complex model. Despite this, the resource usage is within acceptable limits, and power consumption (not directly measured, but inferred from utilization and clock frequency) is expected to be much lower than the GPU's $\sim$225~W TDP, highlighting the efficiency of the FPGA solution.

When compared to GPU-based solutions, the FPGA-based LMIINet implementation demonstrates a significant reduction in power consumption while maintaining competitive performance in terms of accuracy and latency. The ENetHQ model, although faster in terms of FPS, has a considerably lower mIoU (36.8\%) compared to the 45\% mIoU of LMIINet. This underscores the advantage of our approach in achieving better predictive performance without exceeding real-time latency requirements.

\section{Discussion}
\subsection{Design Trade-Offs and Resource Utilization}
The ENetHQ design~\cite{ref2} attains ultra-low latency by pruning and aggressive mixed-precision quantization so that the entire network fits on-chip, yielding low BRAM/LUT/FF usage on ZCU102. Our LMIINet-on-CGRA4ML deliberately spends additional resources on ZCU104 to preserve model capacity and accuracy (\(\sim45\%\)~mIoU vs. 36.8\%), enabled by off-chip streaming and support for depthwise and \(1{\times}1\) convolutions together with simplified attention. This hardware--software co-design maintains real-time throughput while narrowing the accuracy gap to the GPU baseline.

\subsection{Energy Efficiency and Thermal Budget}
In automotive edge settings, power and thermal headroom are primary constraints. While the GPU baseline targets a desktop-class RTX~2080Ti (\(\sim\SI{225}{W}\) TDP), the FPGA design operates within a low device power at \SI{200}{MHz} (inferred from utilization and clocking), supporting fanless or lightly cooled deployments. This order-of-magnitude reduction in power, together with bounded latency, is critical for multi-sensor perception stacks.

\subsection{Implications for Multi-Camera Pipelines}
ENetHQ demonstrates very high aggregate FPS at batch sizes useful for multi-camera ingest, but its accuracy ceiling limits downstream planning performance. Our LMIINet mapping maintains timing for 20~FPS monocular feeds; scaling to multi-camera or stereo can exploit CGRA4ML's reconfiguration to time-multiplex layers while maintaining deterministic per-stream deadlines.

\subsection{Limitations and Opportunities}
Although pixel accuracy is high (\(\sim90\%\)), mIoU (\(\sim45\%\)) indicates room for improvement on small/distant instances---a known challenge in Cityscapes~\cite{ref9}. Adding native support for missing decoder ops (e.g., upsample2d), pursuing higher clock targets, and exploring lightweight instance-aware refinements are promising paths (cf. Future Work).

\section{Discussion}
\subsection{Design Trade-Offs and Resource Utilization}
The ENetHQ design~\cite{ref2} attains ultra-low latency by pruning and aggressive mixed-precision quantization so that the entire network fits on-chip, yielding low BRAM/LUT/FF usage on ZCU102. Our LMIINet-on-CGRA4ML deliberately spends additional resources on ZCU104 to preserve model capacity and accuracy (\(\sim45\%\)~mIoU vs. 36.8\%), enabled by off-chip streaming and support for depthwise and \(1{\times}1\) convolutions together with simplified attention. This hardware--software co-design maintains real-time throughput while narrowing the accuracy gap to the GPU baseline.

\subsection{Energy Efficiency and Thermal Budget}
In automotive edge settings, power and thermal headroom are primary constraints. While the GPU baseline targets a desktop-class RTX~2080Ti (\(\sim\SI{225}{W}\) TDP), the FPGA design operates within a low device power at \SI{200}{MHz} (inferred from utilization and clocking), supporting fanless or lightly cooled deployments. This order-of-magnitude reduction in power, together with bounded latency, is critical for multi-sensor perception stacks.

\subsection{Implications for Multi-Camera Pipelines}
ENetHQ demonstrates very high aggregate FPS at batch sizes useful for multi-camera ingest, but its accuracy ceiling limits downstream planning performance. Our LMIINet mapping maintains timing for 20~FPS monocular feeds; scaling to multi-camera or stereo can exploit CGRA4ML's reconfiguration to time-multiplex layers while maintaining deterministic per-stream deadlines.

\subsection{Limitations and Opportunities}
Although pixel accuracy is high (\(\sim90\%\)), mIoU (\(\sim45\%\)) indicates room for improvement on small/distant instances---a known challenge in Cityscapes~\cite{ref9}. Adding native support for missing decoder ops (e.g., upsample2d), pursuing higher clock targets, and exploring lightweight instance-aware refinements are promising paths (cf. Future Work).

\section{Conclusion}

This work presents a comprehensive approach to real-time semantic segmentation for autonomous vehicles, leveraging the lightweight LMIINet architecture and the CGRA4ML hardware framework for FPGA deployment. Through extensive architectural modifications and hardware-aware optimizations, we successfully mapped LMIINet onto an FPGA platform, achieving a mean Intersection-over-Union (mIoU) of 45\% and a latency of 50~ms per frame. These results demonstrate that the proposed solution meets the stringent requirements of real-time inference and energy efficiency in autonomous driving scenarios.

Key contributions of this work include:
\begin{itemize}
    \item The design and quantization of LMIINet for hardware compatibility, including the integration of efficient encoder-decoder structures, lightweight feature interaction bottlenecks, and transformer-based global context modeling.
    \item The adaptation of the Coarse-Grained Reconfigurable Array for Machine Learning (CGRA4ML) framework to support the deployment of modern neural networks on FPGAs, enabling dynamic reconfiguration and efficient resource utilization.
    \item A thorough evaluation of the FPGA implementation, including training dynamics, simulation results, and comparison with GPU-based solutions, highlighting the advantages in power consumption and hardware utilization.
\end{itemize}

The findings confirm that FPGAs, when paired with optimized neural architectures and frameworks like CGRA4ML, can deliver high-performance, low-latency semantic segmentation suitable for real-world autonomous vehicle applications. This work lays the foundation for further research into scalable, energy-efficient deep learning deployments on edge hardware.

\section{Future Work}

While the current implementation achieves promising results, several avenues remain for future improvement:

\subsection*{Support for Additional Layers}
Block-based operations such as channel shuffle and learnable channel coefficients were not implemented in the current LMIINet architecture. Integrating these features could improve feature extraction and local/global feature interaction, potentially leading to higher segmentation accuracy and better performance on resource-constrained devices.

\subsection*{Increasing Clock Speed}
The Verilator simulation suggests that increasing the clock frequency could further reduce latency and improve real-time performance. Future work should focus on careful frequency tuning to avoid simulation instability and maximize throughput.

\subsection*{Support for Missing Layers in CGRA4ML}
Currently, the Coarse-Grained Reconfigurable Array for Machine Learning (CGRA4ML) framework lacks support for certain layers, such as 2D upsampling (\texttt{upsample2d}), which impacts decoder performance. Adding this functionality would enhance the decoder and overall segmentation quality.

\subsection*{Semantic Segmentation-Specific Library}
Developing a dedicated library for semantic segmentation within CGRA4ML could optimize deployment and improve real-time inference, making the framework more suitable for a wider range of segmentation tasks.

\subsection*{Lighter Model Variants and Hardware Scaling}
Exploring lighter model variants or deploying on larger FPGA chips could enable the use of HLS4ML, which offers better memory access performance and more efficient computation compared to CGRA4ML. This direction could further improve efficiency and scalability for real-world applications.

These improvements will guide the next phase of research, aiming to make LMIINet even more efficient for real-world autonomous driving applications.


\balance

\end{document}